\documentclass{edm_article}
\usepackage{hyperref}
\usepackage{graphicx}
\usepackage{arydshln}  
\usepackage{booktabs}
\begin{document}

\title{Extracting Causal Relations in Deep Knowledge Tracing}

% Submissions for EDM are double-blind: please do not include any author names or affiliations in the submission. 
% Anonymous authors:
\numberofauthors{3}
\author{
    Kevin Hong\\
    \affaddr{University of California, Los Angeles}\\
    \email{kevinhong1167@ucla.edu}
    \and
    Kia Karbasi\\
    \affaddr{University of California, Los Angeles}\\
    \email{kiakarbasi@ucla.edu}
    \and
    Gregory Pottie\\
    \affaddr{University of California, Los Angeles}\\
    \email{pottie@ee.ucla.edu }
}

\maketitle

\begin{abstract}
A longstanding goal in computational educational research is to develop explainable knowledge tracing (KT) models. Deep Knowledge Tracing (DKT), which leverages a Recurrent Neural Network (RNN) to predict student knowledge and performance on exercises, has been proposed as a major advancement over traditional KT methods. Several studies suggest that its performance gains stem from its ability to model bidirectional relationships between different knowledge components (KCs) within a course, enabling the inference of a student’s understanding of one KC from their performance on others. In this paper, we challenge this prevailing explanation and demonstrate that DKT’s strength lies in its implicit ability to model prerequisite relationships as a causal structure, rather than bidirectional relationships. By pruning exercise relation graphs into Directed Acyclic Graphs (DAGs) and training DKT on causal subsets of the Assistments dataset, we show that DKT’s predictive capabilities align strongly with these causal structures. Furthermore, we propose an alternative method for extracting exercise relation DAGs using DKT’s learned representations and provide empirical evidence supporting our claim. Our findings suggest that DKT’s effectiveness is largely driven by its capacity to approximate causal dependencies between KCs rather than simple relational mappings.
\end{abstract}

 % Replace with your own 3-5 keywords

\section{Introduction}
Computer-assisted educational technology, such as intelligent tutoring systems~\cite{fengSystematicReviewLiterature2021, kulikEffectivenessIntelligentTutoring2016}, enables personalizing activities to suit individuals with varying levels of proficiency. A key component of these adaptive systems is knowledge tracing (KT), which aims to estimate students’ mastery of several exercise concepts, known as knowledge components (KCs), as they interact with the corresponding exercises~\cite{explainable_kt, chenPrerequisiteDrivenDeepKnowledge2018}. Formally, \( x_i = \{e_t, a_t\} \) represents a student’s answer pair, where \( e_t \) represents the exercise ID, and \( a_t \) represents whether the student answered correctly or incorrectly. Given a series of past interactions, \( X = \{x_1, x_2, \dots, x_t\} \) and the next concept exercise, \( e_{t+1} \), the task of the KT model is to estimate the likelihood of the student answering correctly, \( a_{t+1} \)~\cite{corbettKnowledgeTracingModeling1994}. Traditionally, Bayesian Knowledge Tracing (BKT)~\cite{corbettKnowledgeTracingModeling1994} was used to perform KT. Despite its popularity, BKT is often critiqued for its binary representation of knowledge states (mastered or not mastered) which oversimplifies the learning process. Several models were later proposed to address these limitations. Learning Factors Analysis ~\cite{cenLearningFactorsAnalysis2006} improved upon BKT by representing learning as a continuous process influenced by multiple exercise interactions and accumulated practice. Performance Factors Analysis ~\cite{jrPerformanceFactorsAnalysis} further captured the complexity of learning by tracking the effects of correct and incorrect prior exercise attempts on performance. These developments laid the groundwork for Deep Knowledge Tracing (DKT) \cite{piechDeepKnowledgeTracing2015}, which popularized a key extension to KT: the ability to implicitly infer relationships between exercises.

The relationships between exercises can take the form of prerequisite dependencies, where understanding one exercise improves performance on another, or corequisites, where exercises depend on each other. Researchers have demonstrated that exercise relationships can be mapped into an exercise graph by analyzing student learning data. An exercise graph represents the dependencies between different concept exercises based on student learning patterns. Creating accurate exercise graphs allows educators to better sequence lessons and ensure that students master foundational concepts before progressing to more advanced ones. KT models are preferred for this task because they learn temporal patterns in student data to capture how mastery of one exercise affects success on another ~\cite{chenPrerequisiteDrivenDeepKnowledge2018, piechDeepKnowledgeTracing2015, tongHGKTIntroducingHierarchical2022}. This enables KT models to uncover learning dependencies that content-based or expert-defined methods may overlook. Furthermore, automating the discovery of exercise relations through the use of KT models offers a scalable alternative to manually defining these relationships. 

The work of DKT made a significant contribution to exercise relation discovery  \cite{piechDeepKnowledgeTracing2015}. Their method assigns an influence score \( J_{ij} \) to every directed pair of exercises \( i \) and \( j \), which is the conditional probability of correctly answering exercise \( j \) after correctly answering exercise \( i \) in the previous timestep, normalized by the sum of such conditional probabilities. The influence score from \( i \) to \( j \) quantifies the prerequisite dependency of the concept \( i \) on the learning concept \( j \). We refer to this method as the DKT method. DKT’s superior performance can be attributed to the model’s ability to learn these influence-based dependencies to estimate student knowledge. Many studies have incorporated exercise relations into subsequent KT models to improve performance and enhance interpretability in the prediction process. For example, Hierarchical Graph Knowledge Tracing (HGKT)~\cite{tongHGKTIntroducingHierarchical2022}, Structure-based Knowledge Tracing (SKT)~\cite{tongStructureBasedKnowledgeTracing2020}, and Deep Knowledge Tracing with Multiple Relations (DKTMR)~\cite{duanMoreAccurateInterpretable2024} are models that track both prerequisite and corequisite relationships between exercises. Graph Attention Knowledge Tracing (GAKT)~\cite{zhaoResearchDeepKnowledge2022} uses a graph attention network to uncover the underlying structure between exercises, learning these relationships using the method introduced in DKT. Graph-based Knowledge Tracing (GKT)~\cite{nakagawaGraphbasedKnowledgeTracing2019} explores multiple statistics-based and learning-based methods to infer the latent graph structure and use the learned graph to perform KT. Among the statistics-based and learning-based methods evaluated, the DKT method to generate graphs performed the best when evaluating performance on the Assistments 2009~\cite{fengAddressingAssessmentChallenge2009} dataset.

The original DKT work introduced influence scores to describe dependencies between exercises. In our work, we build on this idea by reinterpreting these dependencies as a causal structure. We formalize this by pruning exercise relations graphs into Directed Acyclic Graphs (DAGs) to reflect prerequisite relations, and show that DKT's predictive performance improves when trained on data filtered through these causal structures. We also introduce an alternative method for extracting exercise relations that yields accurate and more stable representations of student knowledge and underlying concept dependencies. To facilitate future research on these ideas, we have published our code \footnote{\url{https://github.com/kevinhongca/dkt-causal-relations}}.

% In this work, we explore the effects of causal exercise relationships in KT. We establish a relationship between exercise relation directed acyclic graphs (DAGs) and the predictive accuracy of Deep Knowledge Tracing (DKT) models, and discover that there is a positive relationship between the causal relations learned in the DAGs and KT performance. Additionally, we propose an alternative method for capturing directed exercise relations and demonstrate that this method can more accurately capture the exercise relations that explain DKT’s performance. To facilitate future research on these ideas, we have published our code \footnote{\url{https://anonymous.4open.science/r/dkt-causal-relations-0DB9}}.

\section{Methodology}
We conduct our study using the Assistments datasets, which are among the largest publicly available KT datasets and are widely used as benchmarks for KT models \cite{abdelrahmanKnowledgeTracingSurvey2023, fengAddressingAssessmentChallenge2009, priharExploringCommonTrends2022}. These datasets capture student interactions over extended periods of time and across a wide range of exercises in grade school mathematics. We utilize three datasets: Assistments 2009 \footnote{\url{https://sites.google.com/site/assistmentsdata/home/2009-2010-assistment-data/skill-builder-data-2009-2010}} (\textit{skill builder data corrected collapsed}), Assistments 2012 \footnote{\url{https://sites.google.com/site/assistmentsdata/2012-13-school-data-with-affect}}
 (\textit{2012-2013 data with predictions 4 final}), and Assistments 2017 \footnote{\url{https://sites.google.com/view/assistmentsdatamining/dataset}} (\textit{anonymized full release competition dataset}). The 2009 dataset contains 346,860 exercise attempts from 4,217 students across 123 exercises. The 2012 dataset includes 6,123,270 attempts from 46,674 students across 265 exercises. The 2017 dataset consists of 942,816 attempts from 1,709 students across 102 exercises. The datasets do not contain any personal information.

We use DKT, implemented via the pyKT library \cite{liuPyKTPythonLibrary2023}, to learn the latent exercise relations and help generate the graphs. We begin by training a DKT model on each of the three Assistments datasets. Then, we switch the models to evaluation mode and apply the DKT method. To ensure the resulting graph represents a causal structure, we apply a minimum threshold to the influence scores. This filtering step removes weaker edges that could introduce cycles, allowing us to construct a DAG. Since the distribution of influence scores varies across the three datasets, we select the minimum dataset-specific thresholds that enforce acyclicity. We use a threshold of 0.0107 for Assistments 2009, 0.0051 for Assistments 2012, and 0.0139 for Assistments 2017.

% To ensure the resulting graph captures purely strong prerequisite-dependent relationships, we apply a minimum threshold to filter weak connections and create a DAG.

Using the exercise relation DAG, we create a causal subset of the Assistments dataset by filtering interactions to include only those involving exercises with at least one incoming or outgoing edge. This ensures that the subset consists exclusively of exercises with learned causal relationships, which allows us to isolate and evaluate the influence of these causal structures. We then retrain the DKT model on this subset and evaluate its predictive accuracy. To isolate the impact of causal structure, we generate five random subsets of the dataset with bidirectional relations. Each random subset contains the same number of exercise concepts as its corresponding DAG-based subset, but the exercises are selected randomly rather than based on causal structure. The number of exercise concepts for the subsets was 60 for Assistments 2009, 83 for Assistments 2012, and 17 for Assistments 2017. We then train a separate DKT model on each random subset and compute the mean and standard deviation of their AUC scores. Finally, we compute a z-score to compare the AUC of the DAG-based subset against the distribution of AUC scores from the random subsets across the three Assistments datasets. This allows us to assess the influence of causal relationships on predictive performance. 

Recognizing a potential limitation in DKT’s standard relation extraction method, where influence scores are computed based on a single correct response per exercise, we propose a modified approach that simulates a student repeatedly answering the same type of exercise correctly until the model’s estimated knowledge of that exercise stabilizes. Rather than relying on a one-time response, we feed multiple consecutive correct answers for a given exercise, allowing the model to iteratively update its estimate of the student’s mastery of all the exercises.\

Let \( \hat{K_t} \) represent the student’s estimated knowledge level of an exercise after the student has answered correctly \( t \) times. We continue feeding correct responses until the estimated mastery stabilizes according to the following criterion:

\[
\hat{K}_t = \hat{K}_{t-1}, \quad \forall t \in [t_0 - T, t_0] \text{ s.t. } t_0 \geq T,
\]

where if the model’s estimate does not change for \( T \) consecutive iterations, we stop feeding additional responses and take \( \hat{K_t} \) as the final knowledge estimate. In our experiments, we set \( T = 100 \), chosen heuristically as a reasonable value to allow the knowledge estimates to stabilize. Then, inspired by the method proposed by Piech et al., we use a modified approach,

\begin{equation} \label{eq:Jij}
    J_{ij} = \frac{z(j|i)}{\sum_k z(k|i)}
\end{equation}

where \( z(j|i) \) is the correctness probability assigned to exercise \( j \) on the next timestep given that the student answered exercise \( i \) correctly \( t_0 \) times on the previous \( t_0 \) timesteps. This iterative process provides a more stable estimate of concept relationships by reducing the influence of single-response variations. However, we acknowledge that this approach has practical limitations, which we discuss in Section~\ref{Discussion}.

We evaluate the effectiveness of our modified relation extraction method by repeating the methodology described earlier in this section using Equation \ref{eq:Jij} to generate newly constructed causal graphs: we extract the new DAG-causal subset, retrain DKT on this subset, and compare its AUC scores against those of randomly selected subsets. Note that when extracting the new DAG-causal subset, we again applied dataset-specific thresholds to the influence scores: 0.0129 for Assistments 2009, 0.0067 for Assistments 2012, and 0.0167 for Assistments 2017. By comparing results from both relation extraction methods, we assess whether probing more can yield more accurate causal relationships that help improve KT performance.

% \begin{figure*}
%     \centering
%     \includegraphics[width=0.8\textwidth]{graph1.pdf} % 80% of the two-column width
%     \caption{A sample black and white graphic
%     that needs to span two columns of text.}
%     \Description{Four flies in various orientations next to each other}
% \end{figure*}

% \begin{figure*}[t]
%     \centering
%     \includegraphics[width=0.8\linewidth]{graph.png}  
%     \Description{Exercise Relation Graph for Assistments 2009 Using Our Proposed Methodology}
%     \caption{Assistments 2009 DAG graph of exercise relations using Equation~\eqref{eq:Jij}. Arrow weight indicates prerequisite connection strength. Topic labels are manually added and color coded.}
%     \label{Figure1}
% \end{figure*}

% \begin{figure*}[t]
%     \centering
%     \includegraphics[width=0.8\linewidth]{table1.png}
%     \Description{Table of Exercise IDs and Their Corresponding Names}
%     \caption{Exercise IDs and their corresponding names for Assistments 2009.}
%     \label{Figure2}
% \end{figure*}

\section{Results}

% \begin{table}
% \centering
% \renewcommand{\arraystretch}{1.2}  % Provide more space between table rows, if you prefer
% \caption{AUC results for all Assistments subsets, divided into DAG-based causal subsets (Causal) and random subsets (Random). Causal subsets are derived using DKT's method.}
% \begin{tabular}{cccc} \hline
% Dataset & \textit{AUC} & Exercises (\#) & \( z \)-score \\ \hline
% 2009 Causal & 0.905 & 60 & 3.50 \\
% 2009 Random & $0.859 \pm 0.013$ & 60 & -- \\ \hdashline
% 2012 Causal & 0.727 & 83 & 1.50 \\
% 2012 Random & $0.721 \pm 0.004$ & 83 & -- \\ \hdashline
% 2017 Causal & 0.718 & 17 & 1.02 \\
% 2017 Random & $0.704 \pm 0.011$ & 17 & -- \\
% \hline
% \end{tabular}
% \label{Table1}
% \end{table}

\begin{table}
\centering
\renewcommand{\arraystretch}{1.2}
\caption{AUC results for all Assistments subsets, divided into DAG-based causal subsets (Causal) and random subsets (Random). Causal subsets are derived using DKT's method.}
\begin{tabular}{cccc}
\hline
Dataset & \textit{AUC} & Exercises (\#) & \( z \)-score \\
\hline
2009 Causal & $0.905$ & 60 & 3.50 \\
2009 Random & $0.859 \pm 0.013$ & 60 & -- \\
\hline
2012 Causal & $0.727$ & 83 & 1.50 \\
2012 Random & $0.721 \pm 0.004$ & 83 & -- \\
\hline
2017 Causal & $0.718$ & 17 & 1.02 \\
2017 Random & $0.704 \pm 0.011$ & 17 & -- \\
\hline
\end{tabular}
\label{Table1}
\end{table}

Across all three Assistments datasets, the causal subsets derived from the DKT method \cite{piechDeepKnowledgeTracing2015} consistently achieved higher AUC scores than the mean AUC of randomly selected subsets. See Table \ref{Table1}. This suggests that well-defined causal knowledge dependencies improve a KT model’s ability to trace student learning, and that KT models appear to learn causal relationships more easily than bidirectional ones. Moreover, these findings introduce a potential evaluation metric for knowledge structure graphs: knowledge structures that more accurately capture prerequisite relationships may yield higher AUC scores when their corresponding concepts are used for training.

% \begin{table}
% \centering
% \renewcommand{\arraystretch}{1.2}  % Provide more space between table rows, if you prefer
% \caption{AUC results for all Assistments subsets, divided into DAG-based modified causal subsets (MC) and random subsets (Random). Modified causal subsets are derived using Equation~\eqref{eq:Jij}.}
% \begin{tabular}{cccc} \hline
% Dataset & \textit{AUC} & Exercises (\#) & \( z \)-score \\ \hline
% 2009 MC & 0.875 & 68 & 1.67 \\
% 2009 Random & $0.851 \pm 0.014$ & 68 & -- \\ \hdashline
% 2012 MC & 0.758 & 90 & 5.93 \\
% 2012 Random & $0.721 \pm 0.004$ & 90 & -- \\ \hdashline
% 2017 MC & 0.712 & 59 & 1.85 \\
% 2017 Random & $0.703 \pm 0.005$ & 59 & -- \\
% \hline
% \end{tabular}
% \label{Table2}
% \end{table}

\begin{table}
\centering
\renewcommand{\arraystretch}{1.2}  % Provide more space between table rows, if you prefer
\caption{AUC results for all Assistments subsets, divided into DAG-based modified causal subsets (MC) and random subsets (Random). Modified causal subsets are derived using Equation~\eqref{eq:Jij}.}
\begin{tabular}{cccc}
\hline
Dataset & \textit{AUC} & Exercises (\#) & \( z \)-score \\
\hline
2009 MC & $0.875$ & 68 & 1.67 \\
2009 Random & $0.851 \pm 0.014$ & 68 & -- \\
\hline
2012 MC & $0.758$ & 90 & 5.93 \\
2012 Random & $0.721 \pm 0.004$ & 90 & -- \\
\hline
2017 MC & $0.712$ & 59 & 1.85 \\
2017 Random & $0.703 \pm 0.005$ & 59 & -- \\
\hline
\end{tabular}
\label{Table2}
\end{table}

We then applied our new proposed method to find directed relationships, and as earlier, the new causal subsets consistently outperformed the mean AUC of the randomly selected subsets. See Table \ref{Table2}. To quantitatively compare the performance of the DKT method and our new method, we compute the \( z \)-scores for all causal subsets, defined as:

\[
z = \frac{\text{AUC}_{\text{causal}} - \mu_{\text{random}}}{\sigma_{\text{random}}},
\]

where \( \text{AUC}_{\text{causal}} \) represents the AUC of the causal subset, while \( \mu_{\text{random}} \) and \( \sigma_{\text{random}} \) denote the mean and standard deviation of the AUCs obtained from the random subsets, respectively. We observe that the \( z \)-scores for our new method are higher than those of the original DKT method for two out of the three datasets. Furthermore, the average \( z \)-score is significantly higher for our proposed method. This suggests that our modified approach may yield more accurate representations of underlying knowledge structures.

\begin{figure*}[t]  % Use figure* to span both columns
    \centering
    \begin{minipage}{0.8\textwidth}  % First figure
        \centering
        \includegraphics[width=\linewidth]{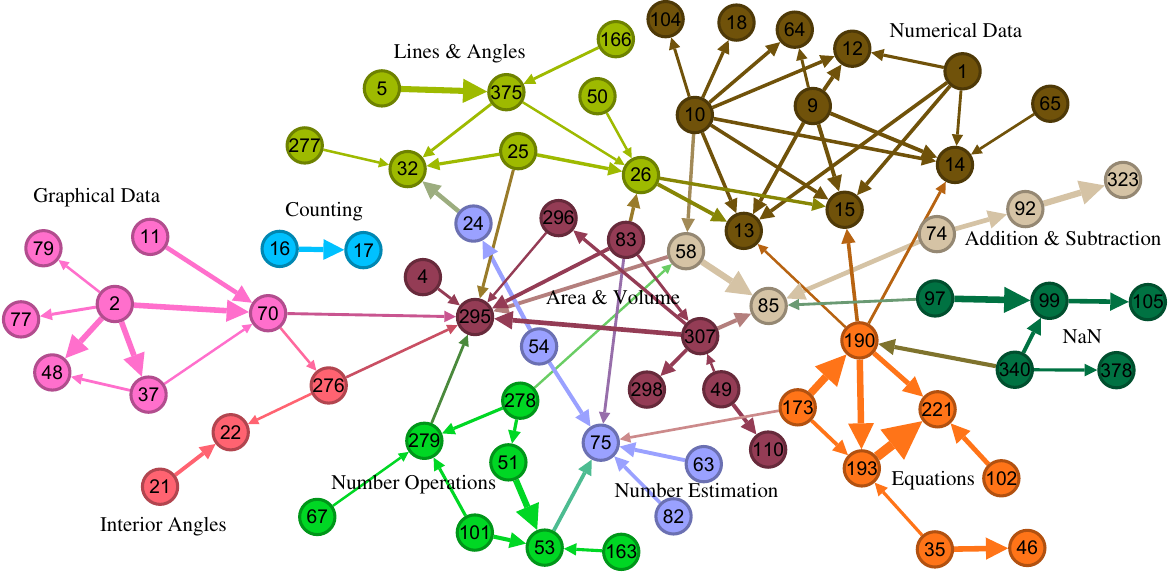}  % Replace with your image
        \Description{Exercise Relation Graph for Assistments 2009 Using Our Proposed Methodology}
    \end{minipage}
    \hfill
    \begin{minipage}{0.8\textwidth}  % Second figure
        \centering
        \includegraphics[width=\linewidth]{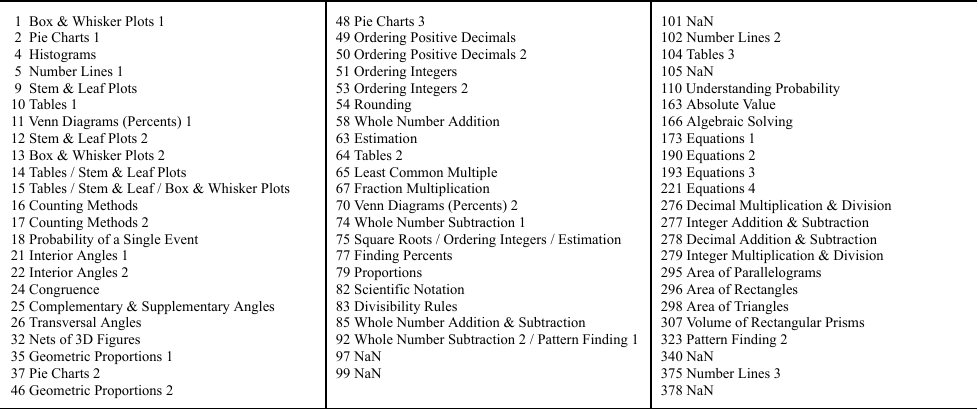}  % Replace with your image
        \Description{Table of Exercise IDs and Their Corresponding Names}
    \end{minipage}
    \caption{Assistments 2009 DAG graph of exercise relations using Equation~\eqref{eq:Jij}. Arrow weight indicates prerequisite connection strength. Topic labels are manually added and color coded.}
    \label{Figure1}
\end{figure*}

To interpret our results further, we use Equation~\eqref{eq:Jij} to generate and analyze the directed exercise relation graphs learned by our models. Figure \ref{Figure1} presents the graph for Assistments 2009, where node numbers correspond to exercise IDs. To better identify the topics between various exercise groups, we apply the algorithms described in \cite{blondelFastUnfoldingCommunities2008, lambiotteLaplacianDynamicsMultiscale2014} to modulate the exercises into topics and use color coding to visually distinguish them. Overall, our approach effectively reveals a meaningful causal structure. Focusing on graphical data topics, we observe that Exercise 2 (Pie Charts 1) serves as a prerequisite for multiple exercises. Since pie charts rely heavily on percentage calculations, it follows that Exercise 2 also contributes to learning related topics, such as Venn Diagrams represented as percentages (Exercise 70) and proportion calculations (Exercise 79). Additionally, the model successfully captures the progression of concepts toward more advanced topics. For example, Exercise 2 (Pie Charts 1) is a prerequisite for Exercise 37 (Pie Charts 2), reinforcing the importance of mastering basic pie chart concepts before progressing. Furthermore, both Exercises 2 and 37 serve as prerequisites for Exercise 48 (Pie Charts 3). While the figure we show provides valuable insights, not all directed edges necessarily indicate true prerequisite relationships, as some connections may result from indirect correlations rather than direct dependencies. Despite these occasional inaccuracies, our method still captures an overall meaningful structure.

\section{Discussion}
\label{Discussion}

In this section, we provide insights into the rationale behind our proposed method for extracting directed relations. We use a DKT model trained on the full Assistments 2009 dataset and set it to evaluation mode to predict the knowledge state of a new simulated student. Our goal is to identify a student's top three highest-ranked exercise concepts, known as KCs, sorted by estimated knowledge mastery, after correctly completing concept 278 (Decimal Addition \& Subtraction). Using the original method, we determine the student’s top three highest predicted KCs immediately after answering concept 278 correctly. In our approach, we allow the student’s understanding of concept 278 to stabilize through one hundred correct responses before extracting the top three highest concept estimates. In other words, we seek three values of \( j \) that maximize \( y(j|i) \) and three that maximize \( z(j|i) \), given that \( i \) = 278 and \( i \) $\neq$ \( j \).

\begin{table*}[ht]
    \centering
    \renewcommand{\arraystretch}{1.2}
    \caption{Comparison between the top three KC masteries between DKT's method and our approach}
    \begin{tabular}{lcc|lcc}
        \toprule
        \multicolumn{3}{c|}{\textit{DKT Method}} & \multicolumn{3}{c}{\textit{Modified Approach}} \\
        \cmidrule(lr){1-3} \cmidrule(lr){4-6}
        Exercise ID & Exercise Name & Mastery & Exercise ID & Exercise Name & Mastery \\
        \midrule
        24 & Congruence & 0.677 & 51 & Ordering Integers & 0.842 \\
        307 & Volume of Rectangular Prisms & 0.673 & 279 & Integer Multiplication \& Division & 0.839 \\
        26 & Transversal Angles & 0.667 & 58 & Whole Number Addition & 0.815 \\
        \bottomrule
    \end{tabular}
    \label{Table3}
\end{table*}

As shown in Table~\ref{Table3}, the top three predicted KC masteries for concept 278 differ between the DKT method and our new approach. The highest-ranked concepts from the DKT method are primarily geometry-related, whereas our approach identifies exercises that are more directly relevant to concept 278, all involving integer operations. We observe that it only takes three consecutive correct responses of exercise 278 for the top three orderings to stabilize. Because \( y(j|i) \) and \( z(j|i) \) correspond to the numerators of the DKT method and our method respectively, these values directly affect how exercise relations are constructed. Referring back to Figure \ref{Figure1}, we see that our method indeed helps relate exercise 278 to the three exercises shown in the Table.

While our modified method may provide a better alternative to the method proposed by DKT, a key limitation of our method is that, in practice, students typically do not answer more than a few questions per exercise. The assumption that a student can be prompted up to 100 times on a given exercise is unlikely to reflect real-world learning scenarios. Future work should experiment with the amount of probing using a more practical number of responses (e.g. 5) or to employ a convergence-based stopping criteria to terminate the iterations when the difference between successive knowledge estimates falls below a certain threshold. 

Another limitation concerns the evaluation setup used to compare DAG-based causal subsets and random subsets. The causal subsets and random subsets may differ in ways beyond their corresponding graph structure. Although we ensure that both subsets contain the same number of exercises, they may still vary in concept coverage, exercise difficulty, or the types of students engaging with the exercises. As a result, some portion of the observed performance gains may be due to differences in exercise or student characteristics, rather than structure alone. Future work should explore more controlled subset construction, such as concept-matched sampling, in which random subsets reflect the distribution of concept types found in the DAG-based subset. For example, if the DAG-based subset consists of 40\% arithmetic, 30\% geometry, and 30\% probability exercises, the corresponding random subset could try to maintain a similar distribution. Strategies like this will help better isolate the impact of causal structure on model performance.

\section{Conclusion}

We show that DKT achieves better predictive performance when trained on DAG exercise subsets, suggesting it effectively learns causal concept dependencies. We introduce a novel method for calculating influence scores that stabilizes knowledge estimates and helps construct accurate exercise relation graphs.
Finally, we acknowledge the limitations regarding the practicality of repeated probing and the need for more controlled experimental comparisons. 

\section{Acknowledgments}

We are grateful to Assistments and pyKT for providing the resources that supported our experiments.

\bibliographystyle{abbrv}
\bibliography{test}  % sigproc.bib is the name of the Bibliography in 

@misc{explainable_kt,
	title = {A {Survey} of {Explainable} {Knowledge} {Tracing}},
	url = {http://arxiv.org/abs/2403.07279},
	doi = {10.48550/arXiv.2403.07279},
	urldate = {2025-03-09},
	publisher = {arXiv},
	author = {Bai, Yanhong and Zhao, Jiabao and Wei, Tingjiang and Cai, Qing and He, Liang},
	month = mar,
	year = {2024},
	note = {arXiv:2403.07279 [cs]},
}

@inproceedings{nakagawaGraphbasedKnowledgeTracing2019,
	address = {Thessaloniki Greece},
	title = {Graph-based {Knowledge} {Tracing}: {Modeling} {Student} {Proficiency} {Using} {Graph} {Neural} {Network}},
	isbn = {978-1-4503-6934-3},
	shorttitle = {Graph-based {Knowledge} {Tracing}},
	url = {https://dl.acm.org/doi/10.1145/3350546.3352513},
	doi = {10.1145/3350546.3352513},
	language = {en},
	urldate = {2025-03-09},
	booktitle = {{IEEE}/{WIC}/{ACM} {International} {Conference} on {Web} {Intelligence}},
	publisher = {ACM},
	author = {Nakagawa, Hiromi and Iwasawa, Yusuke and Matsuo, Yutaka},
	month = oct,
	year = {2019},
	pages = {156--163},
}

@misc{piechDeepKnowledgeTracing2015,
	title = {Deep {Knowledge} {Tracing}},
	url = {http://arxiv.org/abs/1506.05908},
	doi = {10.48550/arXiv.1506.05908},
	urldate = {2025-03-09},
	publisher = {arXiv},
	author = {Piech, Chris and Spencer, Jonathan and Huang, Jonathan and Ganguli, Surya and Sahami, Mehran and Guibas, Leonidas and Sohl-Dickstein, Jascha},
	month = jun,
	year = {2015},
	note = {arXiv:1506.05908 [cs]},
}

@inproceedings{tongStructureBasedKnowledgeTracing2020,
	title = {Structure-{Based} {Knowledge} {Tracing}: {An} {Influence} {Propagation} {View}},
	shorttitle = {Structure-{Based} {Knowledge} {Tracing}},
	url = {https://ieeexplore.ieee.org/document/9338285/},
	doi = {10.1109/ICDM50108.2020.00063},
	urldate = {2025-03-09},
	booktitle = {2020 {IEEE} {International} {Conference} on {Data} {Mining} ({ICDM})},
	author = {Tong, Shiwei and Liu, Qi and Huang, Wei and Hunag, Zhenya and Chen, Enhong and Liu, Chuanren and Ma, Haiping and Wang, Shijin},
	month = nov,
	year = {2020},
	note = {ISSN: 2374-8486},
	pages = {541--550},
}

@misc{tongHGKTIntroducingHierarchical2022,
	title = {{HGKT}: {Introducing} {Hierarchical} {Exercise} {Graph} for {Knowledge} {Tracing}},
	shorttitle = {{HGKT}},
	url = {http://arxiv.org/abs/2006.16915},
	doi = {10.48550/arXiv.2006.16915},
	urldate = {2025-03-09},
	publisher = {arXiv},
	author = {Tong, Hanshuang and Wang, Zhen and Zhou, Yun and Tong, Shiwei and Han, Wenyuan and Liu, Qi},
	month = aug,
	year = {2022},
	note = {arXiv:2006.16915 [cs]},
}

@article{duanMoreAccurateInterpretable2024,
	title = {Towards more accurate and interpretable model: {Fusing} multiple knowledge relations into deep knowledge tracing},
	volume = {243},
	issn = {0957-4174},
	shorttitle = {Towards more accurate and interpretable model},
	url = {https://www.sciencedirect.com/science/article/pii/S0957417423030750},
	doi = {10.1016/j.eswa.2023.122573},
	urldate = {2025-03-09},
	journal = {Expert Systems with Applications},
	author = {Duan, Zhiyi and Dong, Xiaoxiao and Gu, Hengnian and Wu, Xiong and Li, Zhen and Zhou, Dongdai},
	month = jun,
	year = {2024},
	pages = {122573},
}

@inproceedings{zhaoResearchDeepKnowledge2022,
	title = {Research on {Deep} {Knowledge} {Tracing} {Model} {Integrating} {Graph} {Attention} {Network}},
	url = {https://ieeexplore.ieee.org/document/9808616},
	doi = {10.1109/PHM2022-London52454.2022.00074},
	urldate = {2025-03-09},
	booktitle = {2022 {Prognostics} and {Health} {Management} {Conference} ({PHM}-2022 {London})},
	author = {Zhao, Zhongyuan and Liu, Zhaohui and Wang, Bei and Ouyang, Lijun and Wang, Can and Ouyang, Yan},
	month = may,
	year = {2022},
	note = {ISSN: 2166-5656},
	pages = {389--394},
}

@article{kulikEffectivenessIntelligentTutoring2016,
	title = {Effectiveness of {Intelligent} {Tutoring} {Systems}: {A} {Meta}-{Analytic} {Review}},
	volume = {86},
	issn = {0034-6543},
	shorttitle = {Effectiveness of {Intelligent} {Tutoring} {Systems}},
	url = {https://doi.org/10.3102/0034654315581420},
	doi = {10.3102/0034654315581420},
	language = {en},
	number = {1},
	urldate = {2025-03-09},
	journal = {Review of Educational Research},
	author = {Kulik, James A. and Fletcher, J. D.},
	month = mar,
	year = {2016},
	note = {Publisher: American Educational Research Association},
	pages = {42--78},
}

@inproceedings{fengSystematicReviewLiterature2021,
	address = {Lincoln, NE, USA},
	title = {A {Systematic} {Review} of {Literature} on the {Effectiveness} of {Intelligent} {Tutoring} {Systems} in {STEM}},
	copyright = {https://ieeexplore.ieee.org/Xplorehelp/downloads/license-information/IEEE.html},
	isbn = {978-1-6654-3851-3},
	url = {https://ieeexplore.ieee.org/document/9637240/},
	doi = {10.1109/FIE49875.2021.9637240},
	language = {en},
	urldate = {2025-03-09},
	booktitle = {2021 {IEEE} {Frontiers} in {Education} {Conference} ({FIE})},
	publisher = {IEEE},
	author = {Feng, Shi and Magana, Alejandra J. and Kao, Dominic},
	month = oct,
	year = {2021},
	pages = {1--9},
}

@article{corbettKnowledgeTracingModeling1994,
	title = {Knowledge tracing: {Modeling} the acquisition of procedural knowledge},
	volume = {4},
	issn = {1573-1391},
	shorttitle = {Knowledge tracing},
	url = {https://doi.org/10.1007/BF01099821},
	doi = {10.1007/BF01099821},
	language = {en},
	number = {4},
	urldate = {2025-03-09},
	journal = {User Modeling and User-Adapted Interaction},
	author = {Corbett, Albert T. and Anderson, John R.},
	month = dec,
	year = {1994},
	pages = {253--278},
}

@inproceedings{chenPrerequisiteDrivenDeepKnowledge2018,
	address = {Singapore},
	title = {Prerequisite-{Driven} {Deep} {Knowledge} {Tracing}},
	isbn = {978-1-5386-9159-5},
	url = {https://ieeexplore.ieee.org/document/8594828/},
	doi = {10.1109/ICDM.2018.00019},
	language = {en},
	urldate = {2025-03-09},
	booktitle = {2018 {IEEE} {International} {Conference} on {Data} {Mining} ({ICDM})},
	publisher = {IEEE},
	author = {Chen, Penghe and Lu, Yu and Zheng, Vincent W. and Pian, Yang},
	month = nov,
	year = {2018},
	pages = {39--48},
}

@article{jrPerformanceFactorsAnalysis,
	title = {Performance {Factors} {Analysis} – {A} {New} {Alternative} to {Knowledge} {Tracing}},
	language = {en},
	author = {Jr, Philip I PAVLIK and Cen, Hao and Koedinger, Kenneth R},
}

@inproceedings{cenLearningFactorsAnalysis2006,
	address = {Berlin, Heidelberg},
	title = {Learning {Factors} {Analysis} – {A} {General} {Method} for {Cognitive} {Model} {Evaluation} and {Improvement}},
	isbn = {978-3-540-35160-3},
	doi = {10.1007/11774303_17},
	language = {en},
	booktitle = {Intelligent {Tutoring} {Systems}},
	publisher = {Springer},
	author = {Cen, Hao and Koedinger, Kenneth and Junker, Brian},
	editor = {Ikeda, Mitsuru and Ashley, Kevin D. and Chan, Tak-Wai},
	year = {2006},
	pages = {164--175},
}

@article{priharExploringCommonTrends2022,
	title = {Exploring {Common} {Trends} in {Online} {Educational} {Experiments}},
	copyright = {Creative Commons Attribution 4.0 International, Open Access},
	url = {https://zenodo.org/record/6853041},
	doi = {10.5281/ZENODO.6853041},
	language = {en},
	urldate = {2025-03-09},
	author = {Prihar, Ethan and {Manaal Syed} and {Korinn Ostrow} and Shaw, Stacy and Sales, Adam and Heffernan, Neil},
	editor = {Mitrovic, Antonija and Bosch, Nigel},
	month = jul,
	year = {2022},
	note = {Publisher: Zenodo},
}

@article{abdelrahmanKnowledgeTracingSurvey2023,
	title = {Knowledge {Tracing}: {A} {Survey}},
	volume = {55},
	issn = {0360-0300},
	shorttitle = {Knowledge {Tracing}},
	url = {https://dl.acm.org/doi/10.1145/3569576},
	doi = {10.1145/3569576},
	number = {11},
	urldate = {2025-03-09},
	journal = {ACM Comput. Surv.},
	author = {Abdelrahman, Ghodai and Wang, Qing and Nunes, Bernardo},
	month = feb,
	year = {2023},
	pages = {224:1--224:37},
}

@article{fengAddressingAssessmentChallenge2009,
	title = {Addressing the assessment challenge with an online system that tutors as it assesses},
	volume = {19},
	issn = {1573-1391},
	url = {https://doi.org/10.1007/s11257-009-9063-7},
	doi = {10.1007/s11257-009-9063-7},
	language = {en},
	number = {3},
	urldate = {2025-03-09},
	journal = {User Modeling and User-Adapted Interaction},
	author = {Feng, Mingyu and Heffernan, Neil and Koedinger, Kenneth},
	month = aug,
	year = {2009},
	pages = {243--266},
}

@article{blondelFastUnfoldingCommunities2008,
	title = {Fast unfolding of communities in large networks},
	volume = {2008},
	issn = {1742-5468},
	url = {http://arxiv.org/abs/0803.0476},
	doi = {10.1088/1742-5468/2008/10/P10008},
	number = {10},
	urldate = {2025-03-09},
	journal = {Journal of Statistical Mechanics: Theory and Experiment},
	author = {Blondel, Vincent D. and Guillaume, Jean-Loup and Lambiotte, Renaud and Lefebvre, Etienne},
	month = oct,
	year = {2008},
	note = {arXiv:0803.0476 [physics]},
	pages = {P10008},
}

@article{lambiotteLaplacianDynamicsMultiscale2014,
	title = {Laplacian {Dynamics} and {Multiscale} {Modular} {Structure} in {Networks}},
	volume = {1},
	issn = {2327-4697},
	url = {http://arxiv.org/abs/0812.1770},
	doi = {10.1109/TNSE.2015.2391998},
	number = {2},
	urldate = {2025-03-09},
	journal = {IEEE Transactions on Network Science and Engineering},
	author = {Lambiotte, R. and Delvenne, J.-C. and Barahona, M.},
	month = jul,
	year = {2014},
	note = {arXiv:0812.1770 [physics]},
	pages = {76--90},
}

@misc{liuPyKTPythonLibrary2023,
	title = {{pyKT}: {A} {Python} {Library} to {Benchmark} {Deep} {Learning} based {Knowledge} {Tracing} {Models}},
	shorttitle = {{pyKT}},
	url = {http://arxiv.org/abs/2206.11460},
	doi = {10.48550/arXiv.2206.11460},
	urldate = {2025-03-10},
	publisher = {arXiv},
	author = {Liu, Zitao and Liu, Qiongqiong and Chen, Jiahao and Huang, Shuyan and Tang, Jiliang and Luo, Weiqi},
	month = jan,
	year = {2023},
	note = {arXiv:2206.11460 [cs]},
}
%this case

\balancecolumns
% That's all folks!
\end{document}